\documentclass[letterpaper]{article}
\usepackage{aaai20}
\usepackage{times}
\usepackage{helvet}
\usepackage{courier}
\usepackage[hyphens]{url} 
\usepackage{graphicx}
\usepackage{graphicx}
\usepackage{amsmath}
\usepackage{xcolor}
\usepackage{tikz}
\tikzset{>=latex}

\title{Self-Play Learning Without a Reward Metric}
\author{Dan Schmidt,\textsuperscript{\rm 1}
  Nick Moran,\textsuperscript{\rm 1}
  Jonathan S. Rosenfeld,\textsuperscript{\rm 2} \\
  \bf \Large Jonathan Rosenthal,\textsuperscript{\rm 3}
  Jonathan Yedidia\textsuperscript{\rm 1}\\
  \textsuperscript{\rm 1}Analog Devices,
  \textsuperscript{\rm 2}Massachusetts Institute of Technology,
  \textsuperscript{\rm 3}Purdue University \\
  \{daniel.schmidt, nicholas.moran,
  jonathan.yedidia\}@analog.com,
  jonsr@csail.mit.edu,
  rosenth0@purdue.edu}

\begin{document}
\maketitle
\begin{abstract}
  The AlphaZero algorithm for the learning of strategy games via
  self-play, which has produced superhuman ability in the games of Go,
  chess, and shogi, uses a quantitative reward function for game
  outcomes, requiring the users of the algorithm to explicitly balance
  different components of the reward against each other, such as the
  game winner and margin of victory. We present a modification to the
  AlphaZero algorithm that requires only a total ordering over game
  outcomes, obviating the need to perform any quantitative balancing
  of reward components.  We demonstrate that this system learns
  optimal play in a comparable amount of time to AlphaZero on a sample
  game.
\end{abstract}

\section{Introduction}

The AlphaZero \cite{DBLP:journals/corr/abs-1712-01815} algorithm
learns to master a two-player competitive game starting with no
knowledge except for the rules of the game.  As with any sort of
reinforcement learning system, it requires a reward function so that
good outcomes can be distinguished from bad ones.  In the case of
AlphaZero, this is by default a binary-valued function, simply
distinguishing wins from losses; for chess a third intermediate value
is added to represent draws.  Since no differentiation is made between
different sorts of wins, the learner has no explicit incentive to win
more convincingly by scoring more points (in the case of games such as
Go) or by concluding the game in fewer moves (in the case of games
such as chess and shogi). As a result, the trained agent, although
superhuman in most aspects of the game, often makes clearly suboptimal
moves from a human point of view when victory is assured, since this
does not affect the reward received.

In fact, in practice these slack moves can occasionally cause its
advantage to slowly slip away, and the result of the game can change
for the worse, indicating that focusing only on the category of result
can impede generalization.  For these reasons, it is desirable if
possible to introduce a secondary objective (score differential or
number of moves) so that the learner can play optimally from that
perspective throughout the entire game.

The simplest way to add a secondary objective is to modify the reward
function so that more extreme game outcomes are associated with
rewards and penalties that are greater in magnitude. However, this
requires the experimenter to set hyperparameters defining an explicit
quantitative tradeoff between the primary and secondary objectives.

Our variant of the AlphaZero algorithm requires only a total ordering
between game outcomes and produces an agent that attempts to maximize
the rank of the outcomes of its games in that ordering, without the
need for any explicit quantitative balancing factor. We introduce a
method of learning optimal play given this ordering, as well as a
network architecture for predicting outcomes from game states that is
independent of reward function. Because we use only the ordering of
outcomes and not any associated value, we can learn in the absence of
a metric that defines an explicit real-valued distance between
outcomes.

\section{Related Work}

AlphaZero \cite{DBLP:journals/corr/abs-1712-01815} is the ancestor of
all the learning algorithms considered here. It explicitly ignores any
secondary objectives such as score differential. Open-source software
designed to reproduce the AlphaZero algorithm, such as Leela
Zero\footnote{\url{https://zero.sjeng.org}} and Leela Chess
Zero\footnote{\url{https://lczero.org}}, share its limitations in this
respect.

KataGo \cite{DBLP:journals/corr/abs-1902-10565} attempts to improve
upon AlphaZero in various respects in the specific domain of Go,
including dispensing a greater reward for larger wins. The additional
reward as a function of score difference is an ad hoc curve, tuned by
hand, that works well in practice.

The Go engine SAI \cite{DBLP:journals/corr/abs-1905-10863} attempts to
maximize the margin of victory by awarding artificial bonus points to
the losing player, inducing the winning player to increase the actual
score difference to overcome that handicap.

Reward shaping, the practice of engineering a reward function to
improve learning performance, has a long history in the field of
reinforcement learning \cite{ng1999policy}.  Through the lens of this
work, our method may be viewed as a type of automatic and adaptive
reward shaping.

The use of rank-based reward functions is common in the evolution
strategy algorithm and its many variants
\cite{rechenberg1973evolution} \cite{wierstra2014natural}.  Such
rank-based functions are typically used as a way to smooth the reward
landscape, detach learning from hand-engineered payoffs, and provide
scale-invariance of the algorithm with respect to raw reward values.
Rank-based rewards have been used to adapt self-play reinforcement
learning algorithms to single-player tasks, in particular
combinatorial optimization problems \cite{laterre2018ranked}.  Deep
Ordinal Reinforcement Learning \cite{zap2019deep} adapts Q-learning to
use an ordinal reward scale (although without the use of a population
ranking) to induce scale-invariance and reduce the need for manual
reward-shaping.

One can avoid the need to hand-tune the magnitude of the reward function
in reinforcement learning by adaptively normalizing rewards as
learning progresses \cite{van2016learning}.

Our method may be viewed as a form of preference-based reinforcement
learning \cite{wirth2017survey}, in which an agent receives feedback
that indicates the relative utility of two states or actions, rather
than an absolute numerical reward, removing the need for hand-tuning
of reward functions.

\section{Method}

We distinguish between the raw \textbf{result} of a game
(win/loss/draw) and the game's \textbf{outcome}, which contains more
detailed information, such as length of game or margin of victory,
allowing us to perform a finer comparison of two games with the same
result.

\subsection{CDF rewards}

We wish to motivate our learner to win more convincingly without
having to introduce an explicit quantiative tradeoff, such as giving
the winning player a reward of $1 + \gamma m$, where $m$ is the margin
of victory. Removing this explicit secondary reward function reduces
the number of hyperparameters and means that we do not have to tune
$\gamma$ to ensure that the risk-reward ratio involved in pursuit of
larger wins is optimal by some criterion.

Our approach requires only a total ordering over game outcomes. Given
that ordering, we can base the reward solely upon it; at the
conclusion of a game, if a player's outcome is superior to some fraction
$f$ of a corpus of games played by similar (possibly identical)
players, we give that player a reward of $f$.

In effect, we are calculating the cumulative distribution function of
observed outcomes and using it as our reward function.  Naturally,
this CDF of representative outcomes changes over the course of
training, so the reward corresponding to a given outcome changes as
well, but the ordering of rewards continues to match the ordering of
outcomes. (For clarity, we work here with a reward scale from $0$ to
$1$, as that matches the semantics of the CDF, but for convenience
our code uses a scale from $-1$ to $+1$ in order to turn it into a
zero-sum game.)

In practice, a sliding window of some number of generations of the
most recent self-play training game outcomes is maintained, from
which the current CDF of outcomes is computed. Before any games are
played at all, the reward function is a constant, and play is
completely random.

The CDFs of game outcomes in a two-player zero-sum game are specific to
each side and are complements of each other. In a game with a
first-player advantage, a narrow win on the board may well be below
the 50th percentile of outcomes for the first player, and
correspondingly above the 50th percentile of outcomes for the second
player. In practice we maintain a CDF from the first player's point of
view and calculate the second player's reward accordingly.

\begin{figure}
  \includegraphics[width=\columnwidth]{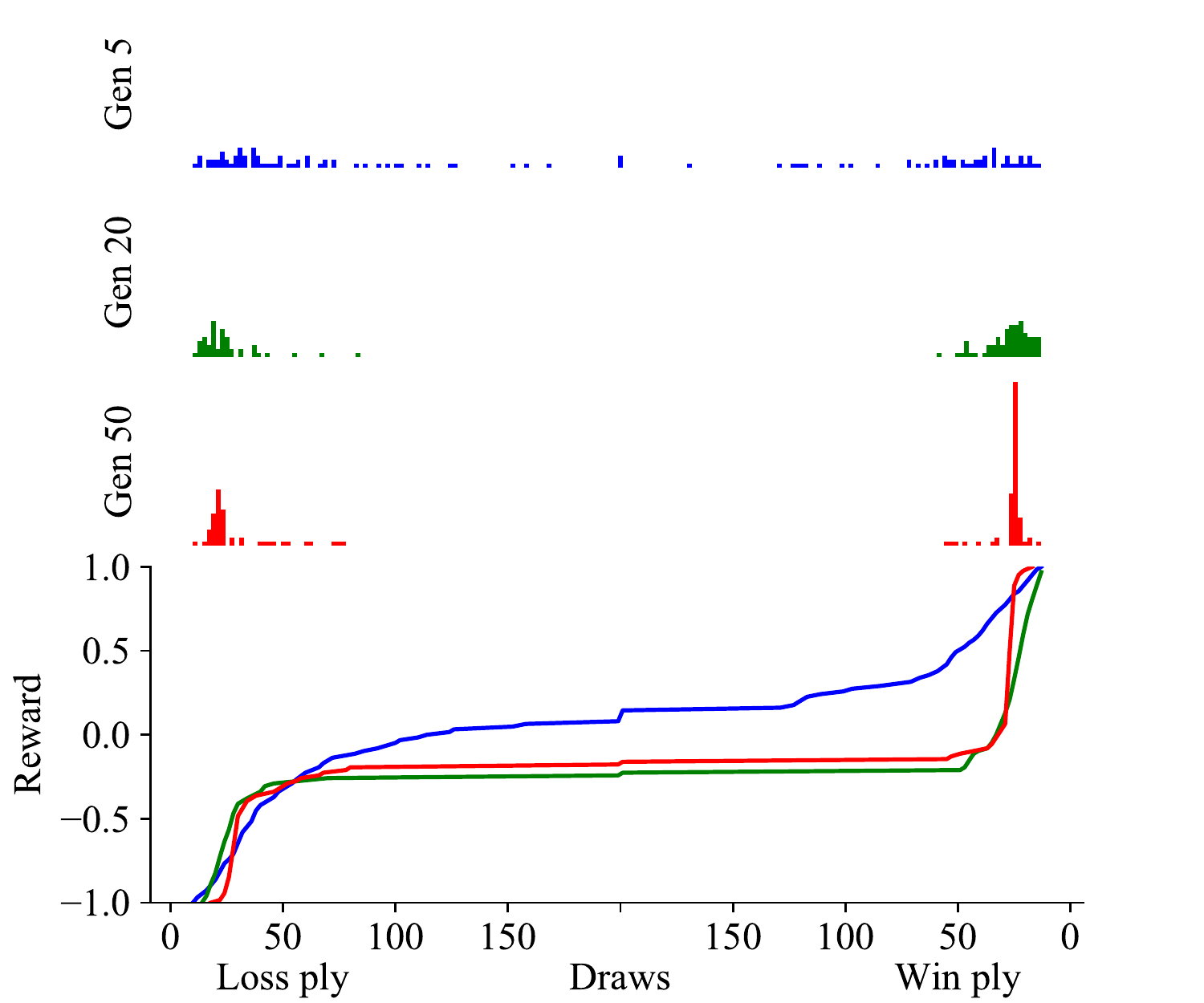}
  \caption{CDF-based rewards at three points during training of a
    $3\times 9$ opposition game agent. The first three plots indicate
    the distribution of recent outcomes after $5$, $20$, and $50$
    generations of self-play; the last plot indicates the resulting
    CDF reward functions.}
  \label{fig:cdf}
\end{figure}

Figure \ref{fig:cdf} illustrates the evolution of a CDF reward
function over the course of training; this is the reward function
induced by the ``CDF/outcome'' agent from Figure
\ref{fig:demerits}. The distribution of observed outcomes becomes more
concentrated as the system learns to play more effectively.  In this
game, two-thirds of the starting positions result in a win for the
first player with optimal play, and it can be seen that this
first-player advantage is recognized by the reward function.

\subsection{Outcome prediction}

In the AlphaZero algorithm, one of the network outputs given a
position is the expected reward of the game that is continued from that
position. In our case, however, the reward corresponding to a game outcome is
not fixed, so we cannot directly predict the expected value of a
position, as the reward function could shift in the future, causing
the value prediction to no longer match the current CDF. Instead, the
network needs to predict a game outcome rather than a value.

This outcome prediction can take many different forms in practice. For
example, if the number of distinct outcomes is limited, the network
can simply produce a categorical output. Our experiments focus on a
game whose secondary objective is to win quickly, and use a game
outcome output head consisting of three categorical outputs (win,
draw, or loss) used as inputs into a softmax function, as well as two
additional predictions for the number of remaining moves in the game
in the separate cases of a win and a loss.  To produce a value, the
rewards corresponding to these three outcomes are computed with the
current CDF, and a final value is computed by weighting these rewards
according to the softmax output. This system captures a reasonable
amount of the uncertainty in the prediction and seems to be a happy
middle ground between a categorical output over a very large number of
outcomes on the one hand and a single point estimate of the outcome on
the other.

As an example, if the post-softmax outputs for win, loss, and draw are
0.60, 0.35, and 0.05 respectively, and the plies-remaining outputs for
wins and losses are 11 and 14 after rounding, the resulting value will
be
$0.60 \cdot F(\text{win in 11}) + 0.35 \cdot F(\text{loss in 14}) +
0.05 \cdot F(\text{draw})$, where $F$ is the current CDF reward
function.

Note that in games where the secondary objective involves the length
of the game, it is most natural for the network to predict the number
of moves remaining (since this is invariant with respect to how many
moves have already been played), rather than to take as input the
number of moves played so far and output the total predicted game
length. Once the number of moves remaining has been predicted, it can
be added to the number of moves already played in order to generate a
final game outcome that is comparable to other ones.

In the case of a game where we supply a larger reward for winning
quickly, there is an additional benefit of outcome prediction over
value prediction. A network that directly predicts the value of a
position would need to take the number of moves already played as an
input and take it into account when calculating a reward, whereas an
outcome-based network can just predict the number of moves remaining,
which can then be combined with the current game state to produce a
value. This can provide an advantage over a value prediction, since
the architecture automatically supplies some generalization of the
reward structure.

Once an outcome, or weighted collection of likely outcomes, has been
generated, it can be dynamically converted to a value using the
current CDF. These values are then used in the standard manner in the
remainder of the MCTS algorithm.

Training of the outcome head is performed by storing the outcome (or
remaining outcome, as described above) in each training example,
rather than reward, and using a suitable loss function, such as
cross-entropy on the win/loss/draw softmax outputs and squared loss on
the relevant remaining-move outputs. For instance, if a training
example is a loss, the output corresponding to the number of remaining
moves in a win has no cost associated with it.

\subsection{Virtual matches and outcome bonuses}

Another way of interpreting these rewards is to imagine that every
game is one half of a two-game match where each player takes each side
once, with a full reward given to the player with the better average
game outcome over the course of the match. To increase our number of
data samples, we can create a large corpus of virtual matches
incorporating every game rather than a single match per game, by
pairing every individual game outcome with every other game
outcome. Given a single game outcome, the expected score of the
virtual matches that result from pairing it with every other game is
exactly the CDF value of that game outcome.

Given this interpretation, we can explore the effects of scoring the
virtual matches differently. For example, we can give the match
winner $1$ point if they win both individual games of the match, but
only $\alpha < 1$ points if they win the match due to having a better
tiebreaker. This effectively creates a discontinuity of size
$1-\alpha$ in the CDF at the boundary between losses and wins, and
recognizes the fact that there is a qualitative perceived difference
between the two categories of outcomes. Of course, this is an
additional hyperparameter.

It is straightforward to create a new reward function that simulates
these virtual matches with bonuses.  Letting $W$ and $L$ be the number
of wins and losses in the corpus, and $i$ being the index in
$[0, L+W-1]$ indicating where the outcome in question lies in the
sorted list of reference outcomes, the associated reward for a loss,
when rescaled to $[-1, +1]$, is

\begin{equation*}
  -1 + \frac{L - \alpha L + 2 \alpha i}{L + W - 1},
\end{equation*}

while the associated reward for a win is

\begin{equation*}
  1 - \frac{W - \alpha W + 2 \alpha (L + W - 1 - i)}{L + W - 1},
\end{equation*}

and the reward for any draw is

\begin{equation*}
\frac{(1 + \alpha)(L - W)}{2(L + W)},
\end{equation*}

which all reduce to the expected formulas when $\alpha = 1$.

This bonus system can be used if some important part of the reward
structure is not captured by the ordering over outcomes. A feature of
the CDF reward system is that it is invariant to reward scaling, but
if the shape of the underlying reward function is important, we can
recognize it using this sort of bonus.

\section{Experiments}

\subsection{The opposition game}

We use the ``opposition game'', a simple game designed to teach
fundamental concepts of chess endgame strategy, as a testbed for our
method because its difficulty is easily parameterizable by changing
the board size, and perfect play is easily achievable, allowing us to
measure the amount of training needed to reach perfection and
quantitatively measure performance during training by comparison to a
perfect player. In addition, from many positions there are multiple
winning moves but some win faster than others, so it is a convenient
testbed for training learners to win quickly. Despite its simplicity,
it will be seen that a straightforward implementation of the AlphaZero
algorithm has difficulty learning to play optimally from a
length-of-game standpoint.

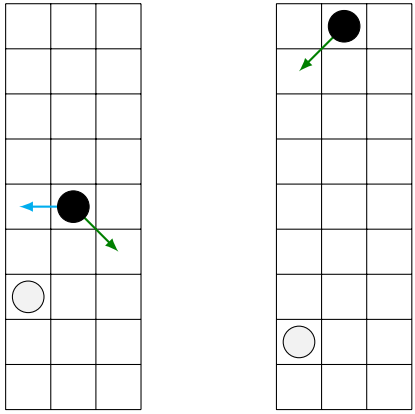
\begin{figure}
  \centering
  \begin{tikzpicture}[scale=0.6]
    \draw (0, 0) grid (3, 9);
    \draw[green!50!black, thick, ->] (1.5, 4.5) -- (2.5, 3.5);
    \draw[cyan, thick, ->] (1.5, 4.5) -- (0.3, 4.5);
    \filldraw[fill=black!05](0.5, 2.5) circle (0.35);
    \filldraw[fill=black](1.5, 4.5) circle (0.35);

    \draw (6, 0) grid (9, 9);
    \draw[green!50!black, thick, ->] (7.5, 8.5) -- (6.5, 7.5);
    \filldraw[fill=black!05](6.5, 1.5) circle (0.35);
    \filldraw[fill=black](7.5, 8.5) circle (0.35);
    
  \end{tikzpicture}
  \caption{Two sample positions from the $3 \times 9$ opposition game
    with Black to play; White is attempting to reach the top rank
    while Black is attempting to reach the bottom. In the left
    position, the dark green move wins in 11 ply, while the light blue
    move wins in 15 ply; all other moves lose. In the right position, 
    the dark green move wins, while all other moves lose.}
  \label{fig:opp}
\end{figure}

The opposition game is played on a $w \times h$ chessboard with two
chess kings, each starting on its own back rank. The players alternate
turns; on one's turn one moves one's own king by one square in any
direction horizontally, vertically, or diagonally. The game is won by
capturing the other player's king (there is no concept of check) or by
reaching the other player's back rank. Two sample positions are
illustrated in Figure \ref{fig:opp}.  It is possible to reach a
position where with perfect play neither player can make progress, but
for any starting position perfect play always ends with a decisive
result.

\begin{figure*}[t]
  \includegraphics[width=\textwidth]{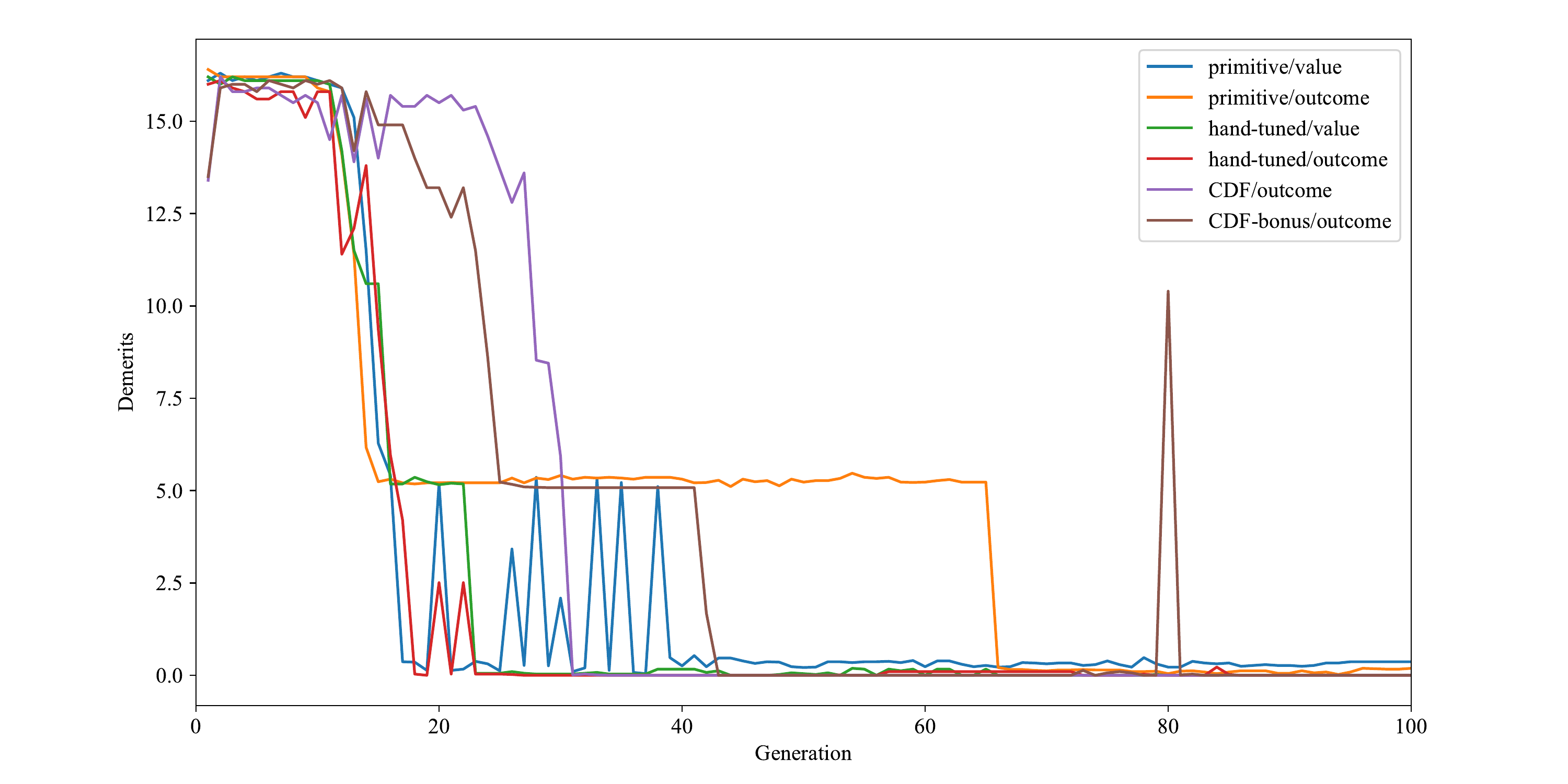}
  \caption{Demerits over time for differently-trained agents. The
    legend indicates the reward function and type of network head for
    each agent. For the CDF-bonus agent, $\alpha = 0.5$. The plateau
    near $5$ demerits is a state where the agent repeatedly makes
    moves that are theoretically winning but delay the end of the
    game, eventually resulting in a timeout against a perfect
    opponent.}
  \label{fig:demerits}
\end{figure*}

Because it is possible to reach a state where neither side can win
with best play, and because a suboptimal player may not be able to
find a win from a theoretically winnable position, we declare a game
to be drawn if a certain number of moves have been played without a
winner. In our experiments this timeout is imposed at a ply value of
$20h$.

We consider all possible initial placements of the kings on their back
ranks, resulting in $w^2$ legal starting positions. Each training game
uses one of these starting positions selected uniformly at random.  It
may appear unprincipled to use a single CDF for multiple starting
positions, since the outcome achieved with optimal play may vary
widely between different starting positions; but the true goal
is to work with a representative distribution of outcomes and so
induce good play, not to compute a mathematically exact result. We
find that using the same CDF for a combination of initial positions
does not impede learning.

\subsection{Results}

The experimental training setup is detailed in Appendix A.

We consider four reward functions for use during training. They are
all zero-sum and range from $-1$ to $+1$, although the extreme values
of this range are not always achievable.

\begin{itemize}
  \item The \textbf{primitive} reward function is just $-1$, $0$, or
    $+1$, depending on the result of the game.
  \item The \textbf{hand-tuned} reward function is a linear
    interpolation from $+1$ for a win ($-1$ for a loss) in zero moves
    to $0$ for a game that has timed out.
  \item The \textbf{CDF} reward function is $2f-1$, where $f$ is the
    fraction of recent training games that have a worse
    outcome for this player's side.
  \item The \textbf{CDF-bonus} reward function treats the CDF function
    as an expected result of virtual matches in which a match win due
    to a tiebreaker (such as game length) is awarded only $\alpha$
    points rather than $1$.
\end{itemize}

The CDF reward functions require us to use a network with an outcome
head rather than a value head, as the mapping from outcome to reward
is dynamic. The other reward functions can be used with either a value
head or an outcome head.

We evaluate the agents by pitting them against a perfect player. After
every generation of training, an $18$-game match is played against the
perfect player, with two games played from each starting position with
kings on their back ranks. A total score for the match is awarded
based on cumulative game score computed with the hand-tuned reward
function. This score is always nonpositive for our agents, since they
are playing against the optimal strategy, so we flip the sign of the
score and refer to it as being in units of demerits. An agent that
lost each of the match games instantaneously (impossible in practice)
would receive $18$ demerits. Figure \ref{fig:demerits} illustrates the
result of this evaluation.

The agents trained with the primitive reward have trouble
achieving a perfect score by this measure; it takes them time to
unlearn suboptimal moves that cause them to time out against the
perfect player, and even at the end of training do not put up optimal
resistance in losing positions, as can be seen by the fact that they
do not converge to $0$ demerits.

The agents trained with the hand-tuned reward reach optimal play
quickly, but require exact knowledge of the specific reward structure
upon which they will be eventually evaluated.

The agent trained with the CDF reward begins with completely random
play, since it cannot distinguish yet between different outcomes.  For
this reason, it requires some additional startup time to generate a
corpus of relevant outcomes, but then quickly learns to maximize its
reward based purely on trying to outperform its peers without the need
for a quantitative metric. The CDF-based agent that is given a winning
bonus learns faster at first, although in this case it took some time
to progress past a suboptimal plateau.

\section{Discussion and Future Work} 

We have presented a variant of the AlphaZero algorithm that does not
require specific numeric rewards to be associated with outcomes but
requires only a total ordering over outcomes. This system can be used
to learn games with a more interesting set of outcomes than just wins,
losses and draws, without having to adjust any hyperparameters
defining some specific relative importance of these additional
factors.

The use of a CDF of game outcomes to define a reward function may
appear problematic at first; after all, with perfect play the CDF of a
nondeterministic game should have all of its weight on a single
outcome. But our CDF is ``softened'' by being assembled from training
games, which are played with Dirichlet noise and a nonzero temperature
(following AlphaZero), so in practice non-optimal outcomes do
occur. In any case, the method requires only an approximate histogram
of typical outcomes so that rough comparisons of the relative degree
of their unusualness can be made. In fact in theory any monotonically
increasing reward function could be used to learn optimal play, but
the CDF reward function has the advantage of having a natural
definition that automatically takes the rarity of different outcomes
into account.

Because the CDF reward function is based on prior self-play games,
play is completely random at the beginning of training and it may
take some time for the games to achieve a terminal state. To jumpstart
the training process, it is possible to begin training with a standard
reward function and then switch to a CDF-based reward function during
training. Since the network predicts outcomes, not values, it is not
made obsolete when the reward function is switched out.

Our experiments have been limited to games where we would like to
induce players to win faster (and lose more slowly), but it is also a
natural fit for score-based games such as Go where the winner would
like to maximize the margin of victory. We would like to explore games
of this nature as well. Now that the ability of this method to learn
has been verified in the more tractable context of the opposition
game, the natural next step is to test it on more complex games that
currently suffer from being trained with a binary or ternary reward
function.

\section{Appendix A: Experimental setup}

We use the same basic learning method as AlphaZero
\cite{DBLP:journals/corr/abs-1712-01815}.  Each generation of training
consists of $25$ self-play games. After each generation of self-play
games, we train our network on the most recent $5$ generations of
games, also using the most recent $5$ generations of games to compute
the CDF used for the reward function.

The network conists of three $3\times 3$ convolutional layers with
$16$ channels and ReLU activations, each followed by a batch norm
layer.  The policy head has one more such convolutional layer with
$32$ channels followed by a single fully-connected layer with a
softmax output. The second head consists of a fully connected layer of
size $64$ with ReLU activation followed by a final fully connected
layer to the output(s).

When this second head is a value head, its output is sent through the
tanh function to produce a value in the range $[-1, +1]$.  When the
second head is an outcome head, it consists of five nodes. Three of
them indicate the relative probability of a win, loss, and draw
result, and are fed into a softmax. Two further nodes predict the
number of ply left in the game in the case of a win and a loss
respectively. Therefore three distinct outcomes (a win of some length,
a loss of some other length, and a draw) with varying probabilities
are proposed, and their corresponding values can be weighted
accordingly.

The CDF reward is determined by the last $5$ generations of self-play
game outcomes. The reward for outcomes that are not present in that
set is calculated by interpolating between the rewards for the two
recorded outcomes that bound it, assuming all possible outcomes are
equally spaced (there are a finite number of these, as we impose a
maximum number of ply on the game). All ply values are scaled by a
factor of $0.1$ in both input and output of the network to keep them
in a reasonable range.

The board representation used for input to the network is specified in
Figure \ref{fig:network-input}.

\begin{figure}
  \begin{center}
    \begin{tabular}{|c|c|}
      \hline Plane & Values \\ \hline
      0 & $1$ at position of Player One, $-\epsilon$ elsewhere \\ 
      1 & $1$ at position of Player Two, $-\epsilon$ elsewhere \\ 
      2 & $\text{ply scale} \cdot \text{current ply} \cdot \epsilon$ \\
      3 & $\epsilon$ if Player One is to move, else $-\epsilon$ \\
      4 & $\epsilon$ (to indicate extent of board to convolution) \\
      \hline
    \end{tabular}
  \end{center}
  \caption{Board representation for the opposition game. The
    placeholder value $\epsilon \equiv \frac{1}{wh}$ is used 
    to normalize the input.}
  \label{fig:network-input}
\end{figure}

The networks are trained on the $5$ most recent generations of
self-play games for $5$ epochs with a constant learning rate of
$0.005$, using SGD with Nesterov momentum of $0.9$.  The loss is a
linear combination of cross-entropy loss on the policy and result
outputs and MSE loss on the value and plies-remaining outputs. For
value heads, these coefficients are $20$ for the policy outputs and
$1$ for the value output; for outcome heads, they are $100$ for the
policy output, $3$ for the three result outputs, and $1$ for the ply
outputs. These coefficients were chosen to roughly balance the losses
from the different categories of output nodes and tuned to work well
in practice.

During self-play, AlphaZero-style Monte Carlo Tree Search is performed
with $20h$ visits, using a temperature of $1.0$ and Dirichlet noise of
$0.5$. At test time, both temperature and Dirichlet noise are set to
$0$.

\bibliography{schmidt}
\bibliographystyle{aaai}
\end{document}